\pdfoutput=1
\documentclass[11pt]{article}
\usepackage{acl}
\usepackage{times}
\usepackage{latexsym}
\usepackage[T1]{fontenc}
\usepackage[utf8]{inputenc}
\usepackage{microtype}
\usepackage{inconsolata}
\usepackage{graphicx}

\usepackage{amsmath}
\usepackage{multicol}
\usepackage{arydshln}
\usepackage{etoolbox,lipsum}
\usepackage{subcaption}
\usepackage{listings}
\usepackage{stfloats}
\usepackage{multirow}
\usepackage{hyperref}
\usepackage{float}
\usepackage{multicol}
\usepackage{color, colortbl}
\usepackage{tcolorbox}
\usepackage{xcolor}
\definecolor{LightCyan}{rgb}{0.93,1,1}
\definecolor{LLightCyan}{rgb}{0.97,1,1}
\definecolor{LightRed}{rgb}{1,0.95,0.95}
\definecolor{LLightRed}{rgb}{1,0.97,0.97}
\definecolor{Gray}{rgb}{0.8,0.8,0.8}
\definecolor{LightGray}{rgb}{0.8,0.8,0.8}
\definecolor{LLightRed}{rgb}{1,0.93,0.95}
\definecolor{LLightBlue}{rgb}{0.9,0.95,1}
\definecolor{LLightGray}{rgb}{0.95,0.95,0.95}
\definecolor{mygreen}{rgb}{0.4,0.7,0.306}
\definecolor{myblue}{rgb}{0.294,0.447,0.796}
\usepackage{hhline}

\definecolor{Gray}{gray}{0.85}
\definecolor{LightCyan}{rgb}{0.88,1,1}

\newcolumntype{r}{>{\columncolor{LLightRed}}c}
\newcolumntype{b}{>{\columncolor{LLightBlue}}c}

\DeclareMathOperator*{\argmin}{arg\,min}

\usepackage[textsize=tiny]{todonotes}

\title{MMR: Evaluating Reading Ability of Large Multimodal Models}

\author{
    Jian Chen\textsuperscript{1}\footnotemark[1],
    Ruiyi Zhang\textsuperscript{2}\footnotemark[2], 
    \textbf{Yufan Zhou\textsuperscript{2},
    Ryan Rossi\textsuperscript{2},}\\
    \textbf{Jiuxiang Gu\textsuperscript{2},
    Changyou Chen\textsuperscript{1}}
    \\ 
    \textsuperscript{1}University at Buffalo~~~~~~~~~
    \textsuperscript{2}Adobe Research
}

\newcommand{\RN}[1]{%
	\textup{\lowercase\expandafter{\it \romannumeral#1}}%
}

\IfFileExists{main_backup.aux}{
  \message{We saw a default backup.aux file, let's use it instead of the main aux file.}
  \nofiles 
  \makeatletter
  \input{main_backup.aux}
  \makeatother
}{}

\begin{document}
\twocolumn[{%
\renewcommand\twocolumn[1][]{#1}%
\maketitle
\vspace{-1.5cm}
\begin{center}
    \captionsetup{type=figure}
    \begin{minipage}[t]{0.42\textwidth}
        \centering
        \includegraphics[width=0.85\linewidth]{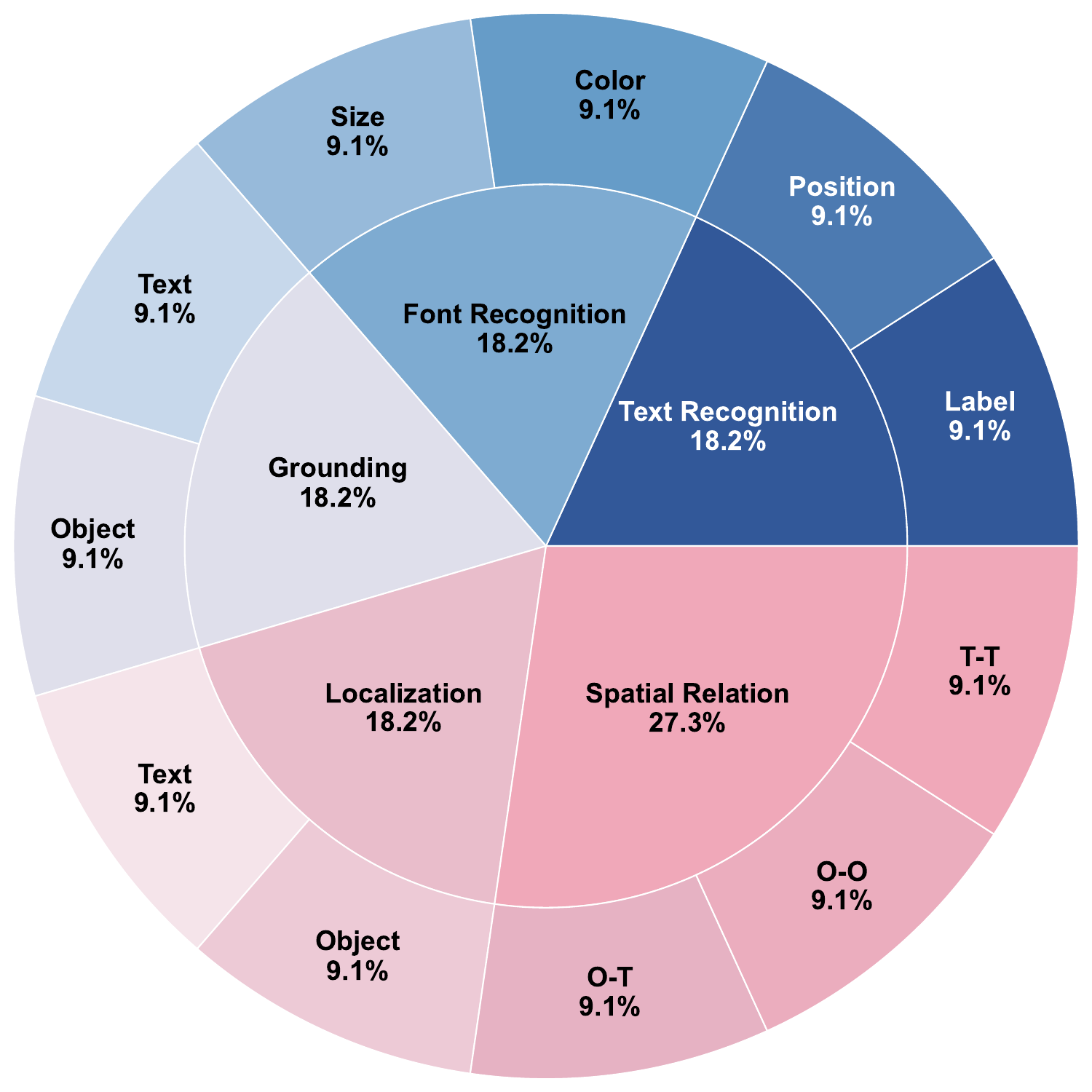}
        \vspace{-0.8em}
        \caption*{(a)}
        \label{fig:pie}
    \end{minipage}
    \hspace{1em}
    \begin{minipage}[t]{0.42\textwidth}
        \centering
        \includegraphics[width=\linewidth]{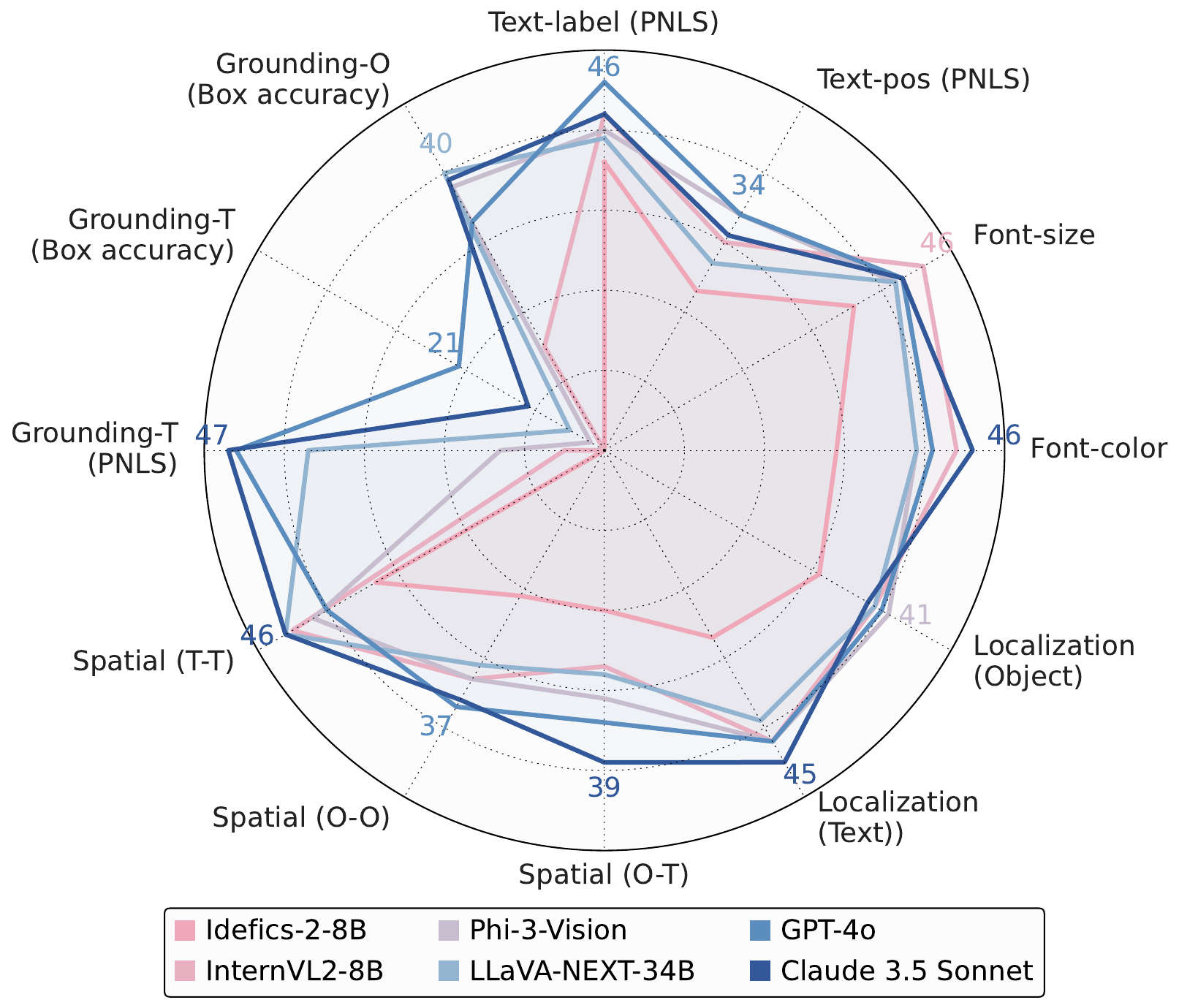}
        \vspace{-2em}
        \caption*{(b)}
        \label{fig:radar}
    \end{minipage}
    \vspace{-1em}
    \caption{(a) Distribution of tasks in MMR Benchmark. (b) Performance of different models on MMR. Existing models show poor text grounding ability and weakness on spatial relationship reasoning.}
     \label{fig:topfigs}
\end{center}%
\vspace{0.5cm}
}]
\renewcommand{\thefootnote}{\fnsymbol{footnote}}
\footnotetext[1]{Work done at University at Buffalo.}
\footnotetext[2]{Corresponding Author}
\renewcommand*{\thefootnote}{\arabic{footnote}}

\begin{abstract}
Large multimodal models (LMMs) have demonstrated impressive capabilities in understanding various types of image, including text-rich images. 
Most existing text-rich image benchmarks are simple extraction-based question answering, and many LMMs now easily achieve high scores. This means that current benchmarks fail to accurately reflect performance of different models, and a natural idea is to build a new benchmark to evaluate their complex reasoning and spatial understanding abilities. In this work, we propose the Multi-Modal Reading (MMR) benchmark\footnote{Project page: \href{https://llavar.github.io/mmr/}{https://llavar.github.io/mmr/}} in 11 diverse tasks to evaluate LMMs for text-rich image understanding. MMR is the first text-rich image benchmark built on human annotations with the help of language models. By evaluating several state-of-the-art LMMs, including GPT-4o, it reveals the limited capabilities of existing LMMs underscoring the value of our benchmark. 
\end{abstract}

\section{Introduction}
Large multimodal models have shown impressive capabilities in understanding various types of image, including text-rich images~\citep{liu2024visual, li2023otter, Li2023LargeMM,zhu2023minigpt4,alayrac2022flamingo,laurenccon2024matters, mckinzie2024mm1}. Existing text-rich image datasets and benchmarks are composed of single-page document images~\cite{mathew2020docvqa,mishra2019ocrvqa,mathew2022infographicvqa} or natural images with scene texts~\cite{textvqa,sidorov2020textcaps}. The questions associated with these datasets typically require simple extraction rather than advanced reasoning or spatial understanding. With LMMs showing significant performance gains~\cite{liu2024llavanext,liu2024textmonkey}, existing benchmarks have almost been solved, as evidenced by the high metric scores. This progress has made it challenging to accurately gauge the true capabilities and differentiate the performance levels of various models.

In this work, we introduce a novel Multi-Modal Reading (MMR) benchmark designed to provide a more rigorous assessment of Language Multimodal Models (LMMs) in the context of text-rich image comprehension. Specifically, MMR is built on the LAION dataset and selectively retains images that exhibit a significant presence of text. We ask human annotators to create comprehensive captions that encompass text content, visual elements, layout structures, and their inherent attributes. Leveraging advanced language models, such as GPT-4V, we generate a challenging Visual Question Answering (VQA) benchamrk. This benchmark surpasses the complexity of existing text-rich image VQA datasets and encompasses 11 distinct tasks. For each of these tasks, we have refined the metrics to better suit the evaluation of LMMs. 

Our comprehensive evaluation of open-source and proprietary models, varying in size, reveals the current limitations of LMMs. Specifically, we still see a gap between open-source and proprietary models. These open-source models usually do not follow the instructions provided and output in the desired format, mainly due to the limited size of the instruction finetuning dataset. LLaVA-Next-34B~\cite{liu2024llavanext} shows the best performance in the object grounding task. Phi-3-Vision~\cite{abdin2024phi} shows impressive performance even with compact size, further demonstrating the importance of data quality. These observations further show that open-source models can perform better than proprietary models in specific skills. All models show poor performance on text grounding tasks, which is an important skill to improve for future LMMs. We anticipate that the MMR benchmark can provide valuable insights for the research community and encourage further advances in the nuanced field of complex visual text understanding.

\section{Related Work}
\label{appendix:related}
\paragraph{Classical VQA Benchmarks}
TextCap~\citep{sidorov2020textcaps} is the first text-rich image captioning dataset.
Text-OCR~\citep{textocr} aims to comprehend text in the context of an image, which is similar to our motivation, but focuses more on text recognition in images rather than understanding. ST-VQA~\citep{STVQA} uses spatial and textual information to answer visually grounded questions, effectively integrating visual and textual cues. OCR-VQA~\citep{mishra2019ocrvqa} focuses on incorporating optical character recognition (OCR) into visual question answering (VQA), which operates primarily on text within images. TextVQA~\citep{textvqa} also takes advantage of the textual information present in the images to answer questions, but with an emphasis on open questions. DocVQA~\citep{mathew2021docvqa} takes this one step further by applying VQA to document images, handling a variety of layouts and formats. InfoVQA~\citep{mathew2022infographicvqa} and ChartQA~\citep{masry2022chartqa} focus on specific subdomains and aim to answer questions about information graphics and chart images, respectively.  All of these benchmarks are mostly composed of extractive questions, while MMR provides complex reasoning evaluations for multimodal LLMs. 

\paragraph{Large Multimodal Model Benchmarks}
Recent advancements in large multimodal models have led to the development of various benchmarks aimed at evaluating their capabilities across different tasks \cite{fu2023mme, li2023seed}. 
MMBench \cite{liu2023mmbench} and MM-Vet~\cite{yu2023mm} offer comprehensive assessments of multimodal model efficacy on recognition-based tasks. More recently, the BLINK \cite{fu2024blink} benchmark was proposed for evaluating a model's nuanced perception abilities beyond recognition. \citet{tong2024cambrian} presents CV-Bench, which adapts existing vision benchmarks \cite{brazil2023omni3d, lin2015microsoft, zhou2019semantic} to formulate natural language questions aimed at testing the spatial comprehension abilities of models. MM-UPD Bench \cite{miyai2024unsolvable} on the other hand, tests a model's ability to recognize and refrain from answering unsolvable VQA problems in the multiple-choice setting. Benchmarks beyond single-image understanding have been designed to assess the ability to understand multiple images~\cite{li2023fine,wang2024mementos}. 

Benchmarks have also been proposed to evaluate more specific abilities of multimodal models. MathVista \citep{lu2023mathvista} focus on mathematical reasoning and SciFIBench \citep{roberts2024scifibench} focused on scientific figure interpretation. Recent work introduced the MMMU benchmark \citep{yue2024mmmu} that provides multi-discipline tasks for evaluation of large multimodal models, that require college-level knowledge about specific subject matters and deliberate reasoning capabilities. Multipanel VQA \cite{fan2024muffin} introduced the multipanel visual question answering task, which involves interpreting multiple image panels arranged as a layout in a single image, such as posters and website screenshots. 
VisualWebBench~\cite{liu2024visualwebbench} assesses the capabilities of LLMs across a variety of web tasks.
MuirBench \cite{wang2024muirbench} is designed to assess the ability to comprehend multiple images simultaneously. There have also been recent work focused on benchmarking multi-modal LLMs for video analysis called Video-MME \citep{fu2024video}. These benchmarks collectively push the boundaries of what large multimodal models can achieve, fostering continuous improvement and innovation. 

\section{Multimodal Reading Benchmark}
For existing text-rich image benchmarks, most of them~\citep{textvqa,textocr,mathew2020docvqa} focus on information extraction, such as DocVQA~\citep{mathew2021docvqa} and TextVQA~\citep{textvqa}. The recently released OCRBench creates a new benchmark by carefully selecting questions from existing benchmarks, and MT-VQA~\cite{tang2024mtvqa} expands multilingual VQA pairs in text-rich images.  To address this problem and provide a better evaluation of LMMs in text-rich images, we propose the MMR benchmark to evaluate multimodal reading ability, including spatial understanding, text recognition, and complex reasoning. Figure \ref{fig:topfigs}(a) shows the distribution of different tasks in MMR and \ref{fig:topfigs}(b) shows the performance of representative models. 

\begin{figure}[!h]
    \centering
    \includegraphics[width=1\linewidth]{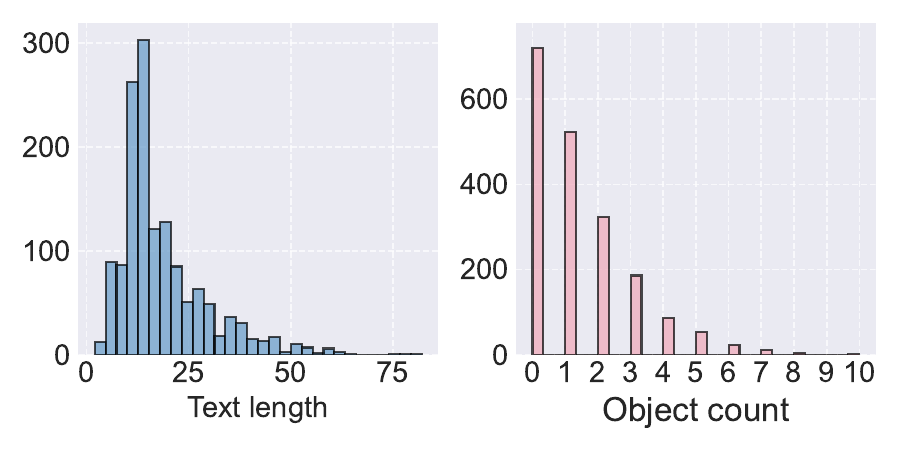}
    \vspace{-2em}
    \caption{Text and Object length distribution in MMR.}
    \label{fig:hist}
    \vspace{-0.5em}
\end{figure}
\begin{figure}[!h]
    \centering
    \includegraphics[width=\linewidth]{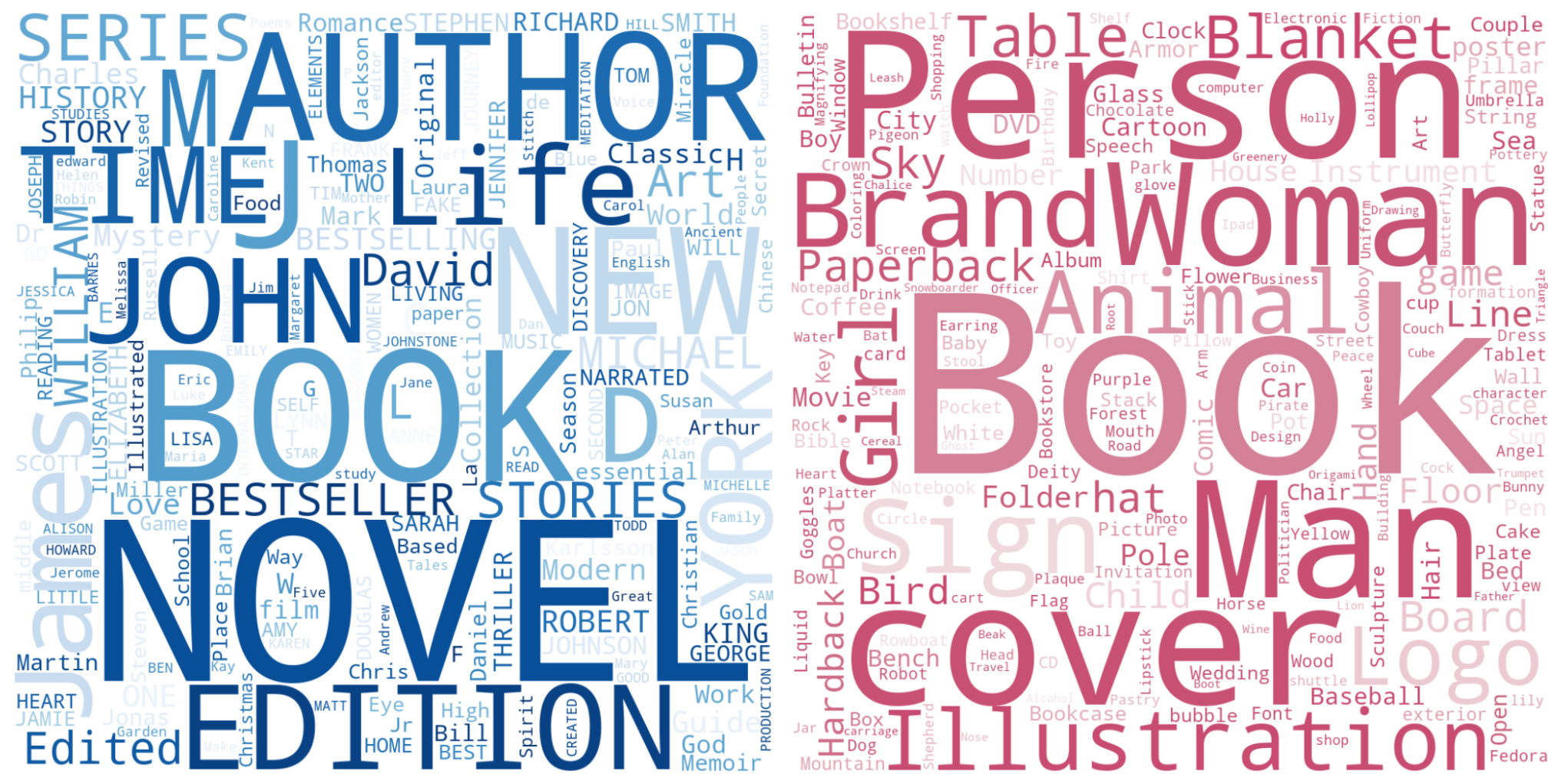}
    \vspace{-1.9em}
    \caption{Wordcloud of text (Left) and object tags (Right) of MMR Benchmark.}
    \label{fig:cloud}
    \vspace{-1.0em}
\end{figure}
\subsection{Statistics}
\begin{figure*}[t]
    \centering
    \includegraphics[width=1\textwidth]{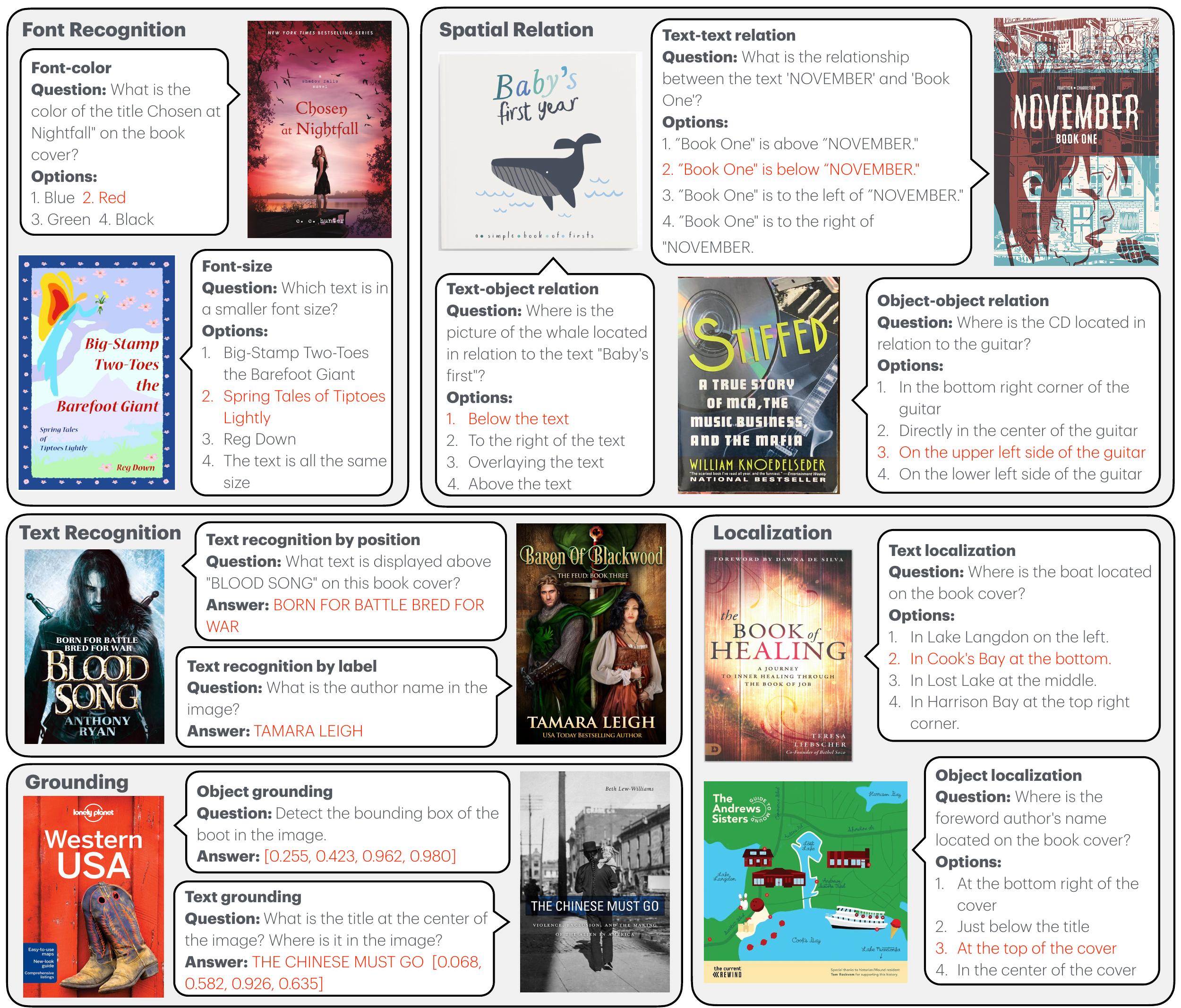}
    \caption{Example questions from MMR to evaluate reading capabilities.}
    \label{fig:demo}
    \vspace{-1em}
\end{figure*}

MMR benchmark comprises pairs of 11 visual question answering tasks on text-rich images. The dataset is constructed based on 408 images selected from a total of 1,931 text-rich images. We included example questions for each task in Figure \ref{fig:demo} for better understanding. 
We categorized all questions into five main classes, which are further divided into 11 more specific types. To ensure a fair evaluation, we manually curated 50 question-answer pairs for each type of question. 

Due to the high diversity of objects and visual text in the benchmark, we do not classify them into a predefined set of labels. Instead, we use an OCR model to detect text, and RAM++ \citep{huang2023open} to generate open-set object tags and display rough content distribution using word clouds. 
Figure \ref{fig:hist} shows the distribution of OCR words and the number of objects detected on the MMR benchmark, with means of 18.45 and 1.35. Figure \ref{fig:cloud} shows the word cloud of text and object tags. 

Considering the poor performance of different models on the text grounding tasks shown in Figure \ref{fig:topfigs}, we create an additional 900 question-answer pairs in text grounding for a comprehensive evaluation and the development of new methods. 

\begin{figure*}[!h]
    \centering
    \includegraphics[width=\linewidth]{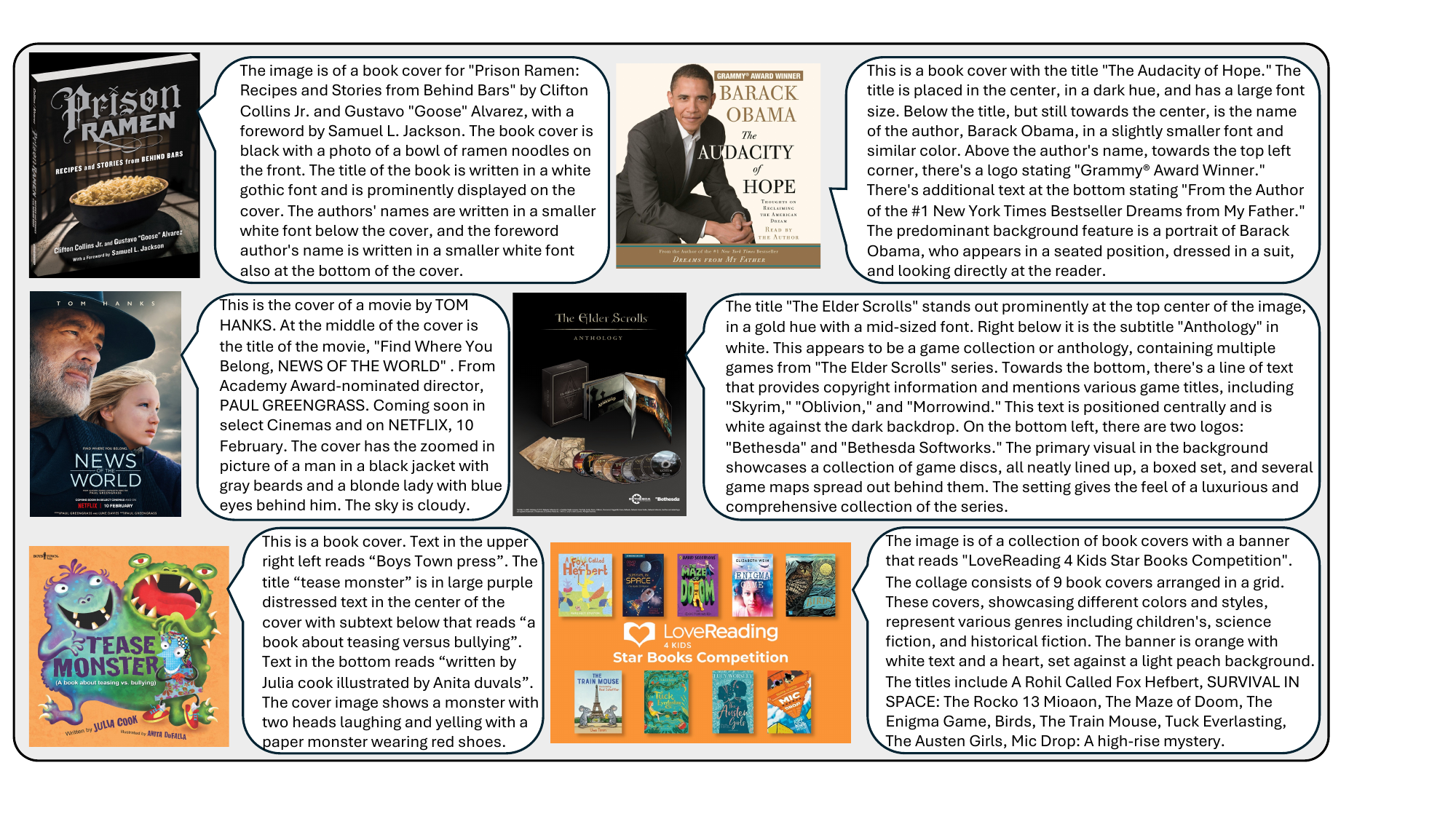}
    \vspace{-2.0em}
    \caption{Examples of human annotated dense captions. All text elements are annotated in detail, such as color, position, and contents. Detailed descriptions of visual elements and layout information are provided as well.}
    \label{fig:humancaption}
    \vspace{-0.6em}
\end{figure*}

\subsection{Data collection and Human Annotations}
We first build a dataset of text-rich images based on LAION-5B~\footnote{\url{https://huggingface.co/datasets/laion/laion-high-resolution}} \citep{schuhmann2022laion} with carefully designed heuristics and machine learning models. Then we ask annotators to give detailed captions for each text-rich image. In addition, we design various prompts and use human annotations to help GPT-4V \citep{yang2023dawn} generate question-answer pairs. After that, human annotators are asked to verify these QA pairs and make corrections as benchmarks.

\paragraph{Machine-Assisted Image Selection}
We filter and maintain text-rich images from LAION-5B dataset. To differentiate between text-intensive document images and natural images, we first compile a binary classifier, a DiT \citep{2022DIT} base model, which was further refined using the RVL-CDIP dataset \citep{harley2015evaluation}, to determine the presence of text in an image. Then we use PaddleOCR to extract all words from the selected images and keep images with more than 20 words and less than 100 words, which eliminates most text-intensive document images.
The final step uses semantic information to select the desired images. A random sample of 20,000 images from the filtered LAION-5B is clustered into 50 groups based on CLIP-ViT-B/32 visual features. After inspecting the clustering results, two clusters are chosen as text-rich images. This cluster model then serves as the filtering mechanism for collecting images that comprise the MMR dataset.

\paragraph{Human Annotated Dense Captions} 
Following the scheme of human-annotated captions from TRINS~\citep{trins2024}, we provide comprehensive annotation instructions and examples to annotators and ask them to (\RN{1}) provide detailed descriptions of visual components, and (\RN{2}) describe the location, attributes, and exact words of the texts in annotations. Our goal is to better translate a text-rich image into text descriptions with minimum information loss. Considering the unstable ability of multimodal understanding and great performance on text-only tasks, this process can provide a reliable source for question-answer pairs generation. Figure \ref{fig:humancaption} shows examples of annotated examples.

\paragraph{Human-Machine Hybrid QA Annotations}
We provide image annotations, including human annotations, OCR results, object detection results, and the images themselves, to GPT-4V to construct QA pairs that test the visual understanding abilities of the vision-language model on text-rich images. These questions are divided into two categories: reading and understanding visual text, and spatial position detection and understanding of visual elements (objects and text). For text recognition and position detection, we prompt the model to output the detected text and bounding-box coordinates in a fixed format. For more complex questions, we create multiple choice questions and require the model to output the index of the correct option to facilitate quantitative evaluation. 

When constructing questions, we use the Azure OCR tool~\footnote{\url{https://learn.microsoft.com/en-us/azure/ai-services/computer-vision/how-to/call-read-api}} to recognize visual text and detect their bounding boxes, and we use Grounding DINO \citep{liu2023grounding} to detect objects. These results serve as the ground truth for certain questions and are provided to GPT-4V to enhance the reliability of the answers to generated questions. All question-answer pairs are manually inspected for accuracy. 

\subsection{Benchmark Tasks}
MMR benchmark encompasses 11 distinct tasks in texts, fonts, visual elements, bounding boxes, spatial relations, and grounding, as demonstrated in Figure \ref{fig:demo}. These tasks can be categorized into text recognition, spatial relationships, localization, and grounding, and all are essential skills to evaluate the reading ability of large multimodal models.

\paragraph{Text Recognition}
For text recognition, we ask a model to retrieve text for a given label, such as title or author name. This requires a model to extract text for a specified label. We also ask a model to retrieve text strings based on their position on the canvas or relative to other elements, testing its localized reading ability. 
\vspace{-0.2em}
\paragraph{Font Recognition}
We use multiple-choice questions about font size and text color to assess the model’s ability of visual text understanding. These questions are created based on OCR results, human annotations, and the image itself. They ask about the color of a specified text and compare the font sizes between two texts within the given image.
\vspace{-0.2em}
\paragraph{Element Localization}
We construct multiple-choice questions for both text and object localization, asking the model to choose the correct region for the target element, such as the bottom or top left corner of the image. This task aims to roughly localize the target text without requiring complicated format requirements for the output.
\vspace{-0.2em}
\paragraph{Spatial Relationship Understanding}
We construct multiple-choice questions to test the model's ability to comprehend pairwise spatial relationships between elements. The questions are categorized into three types: object-text, object-object, and text-text pairs. These questions are generated based on images, human annotations, element bounding boxes, and optionally OCR results.
\vspace{-0.2em}
\paragraph{Object Grounding}
We prompt the model to output relative bounding-box coordinates within a range of 0 to 1 in a Python list format, which requires high object localization precision and instruction-following ability. The ground truth is generated by combining human and model-based annotations.
\vspace{-0.2em}
\paragraph{Text Grounding}
We construct text grounding questions that require a model to output both the text string and the bounding box coordinates simultaneously, in a specified format that concatenates a string with a Python list for auto-extraction. The target text is specified by its rough location on the 3x3 grid and the corresponding text labels. This task demands high text localization precision and even higher instruction-following ability compared to object grounding, which only asks for box coordinates.
\subsection{Quality Control} 
To ensure the correctness and quality of the QA pairs in our data set, we combine human- and model-based annotations to select ideal images for each task. We then manually curate 50 questions per task to assemble a high-quality QA dataset.

\paragraph{Bounding Box} 
For text grounding and recognition, we extract text content from human annotations~\cite{trins2024} and match it with OCR results, discarding images where OCR fails to capture all annotated texts. The OCR model provides quadrilateral bounding boxes, which we filter using a 30-degree threshold on the average horizontal tilt angle of the upper and lower edges to exclude sloping text unsuitable for rectangular box detection. For object grounding, we utilize RAM++ \citep{huang2023open} to exclude images without objects. We then manually label object tags and feed these tags into Grounding-DINO to detect rectangular bounding boxes for specified objects. 

\paragraph{Spatial Location}
We extract the center coordinates of each bounding box and assign a rough position using a 3x3 grid on the canvas. This rough positioning is used as a condition for text recognition and text grounding tasks.

\paragraph{Multiple Choices}
For multiple-choice questions, we prompt GPT-4V to generate a question with four choices and the true answer in one response. Multiple choice provides an easy way to perform automatic evaluation, avoiding the usage of hard-matching metric scores~\cite{papineni2002bleu,lin2004rouge} or the involvement of large models in the evaluation process~\cite{liu2023g}.

\paragraph{Human Verification}
We ask human annotators to review the MMR benchamrk. Both questions and answers are manually checked, and annotators will correct the answer or rewrite the question if it does not belong to the target task.

\section{Experimental Results}
We evaluate the performance of popular vision-language models on the MMR benchmark. Our assessment includes seven open-source vision-language models of different sizes: Monkey-Chat \citep{li2023monkey}, Idefics \citep{laurenccon2024obelics}, Idefics-2 \citep{laurenccon2024matters}, LLaVA-v1.5 \citep{liu2024visual}, LLaVA-NEXT \citep{liu2024llavanext}, Phi-3-Vision \citep{abdin2024phi}, and InternVL2 \citep{chen2023internvl}. Additionally, we evaluate five proprietary vision-language models: Qwen-vl-plus, Qwen-vl-max \citep{Qwen-VL}, Claude 3.5 Sonnet \citep{anthropic2024a}, GPT-4V \citep{yang2023dawn}, and GPT-4o. We include the prompts used for our experiment in Appendix \ref{appendix:prompt}. All experiments were performed on a single A100-80GB GPU.

\subsection{Evaluation metrics}
\label{sec:metrics}
We evaluate the model's performance on all tasks using three metrics tailored to the output type. For multiple-choice questions, performance is measured by the number of correct choices made by the models. For the grounding task, we assess the quality of detected bounding boxes using the IoU score, and for text recognition, we propose a new metric, PNLS, to compare text strings. To facilitate the computation of a total score across all tasks for comparing the overall performance of models, we convert the continuous IoU and PNLS metrics to binary scores using a threshold. In our benchmark, we set the thresholds to 0.3 for IoU and 0.9 for PNLS, respectively. The two metrics are explained as follows:

\vspace{-0.3em}
\paragraph{PNLS}
For text recognition tasks, we propose Partial Normalized Levenshtein Similarity (PNLS), a variant of normalized levenshtein similarity (NLS) \citep{biten2019scene}. PNLS adapts the global alignment algorithm into a local-global version \citep{sellers1980theory}, which avoids penalizing extra prefix or suffix characters. This makes it more effective for evaluating text recognition results from language models, as these models often produce verbose outputs to improve user experience.

Compared to the normalized Levenshtein similarity (NLS), PNLS uses the length of the region aligned with the true text string as the normalization factor. This aligned region is determined through dynamic programming. The score still ranges from 0 to 1 and positively correlates with performance. The motivation behind this design is to avoid penalizing extra prefixes or suffixes in a model's output. AccANLS \citep{zhang2024exploring} was proposed for the same purpose. However, it only spares penalties on prefixes and suffixes when there is an exact match of the true text string in the model's output.

The PNLS metric is formally defined as follows: String $\mathcal{T}_{1,m}=t_1\dots t_m$ represents the true answer and $\mathcal{S}_{1,n}=s_1\dots s_n$ is a model generated string. We first identify the sub-string of $\mathcal{S}$ that has the minimum edit distance to $\mathcal{T}$. Specifically, we first construct a scoring matrix $\mathbf{F}$ of size $(m+1)\times(n+1)$, where $F_{i,j}$ stores the smallest edit distance between the $i$-prefix ${\mathcal T}_{1,i}$ and any sub-string ${\mathcal S}_{x, j}$, $\forall x \in \{1,\dots,j-1\}$ that ends at position $j$. The scoring matrix can be computed recursively
\begin{equation*}
\label{recur}
\small
    F_{i,j} = \left\{
                \begin{array}{ll}
                  0 &\text{if } i = 0 \\
                  m &\text{if } j = 0\\
                  \min \left (
                \begin{array}{l}
                F_{i-1,j-1} + c(t_i,s_j)\\
                F_{i-1,j} + 1\\
                F_{i,j-1} + 1 
                \end{array}
                \right )
                &\text{otherwise},\\
                \end{array}
              \right.
\end{equation*}
where $c$ is the substitution cost that takes a value of $0$ if $t_i = s_j$ and $1$ otherwise. Once ${\bf F}$ is computed, the minimum value in the last row is the optimal edit distance and the end index of the matched sub-string $j'=\argmin_{j}(F_{m+1,j})$. The start index $i'$ can be found by tracing back the the computation of Eq.(\ref{recur}). Finally, the PNLS is computed as: $m / (m + j'-i'+1)$.

\vspace{-0.3em}
\paragraph{IoU Scores}
For object and text grounding tasks, we use the mean Intersection over Union (IoU) score to evaluate the model's accuracy. We also report the number of valid outputs that follow the required format, evaluating the instruction following ability, and allowing a script to automatically extract the coordinates and text strings. 

\begin{table*}[!ht]
\centering
\fontsize{9}{13}\selectfont
\setlength\arrayrulewidth{0.6pt}
\resizebox{\textwidth}{!}{
\setlength{\tabcolsep}{4pt}
\begin{tabular}{lc|bb|cc|cc|ccc|r|br|c}
\hline
\rowcolor{LLightGray} &  & \multicolumn{2}{c}{\textbf{Text}} & \multicolumn{2}{c}{\textbf{Font}} & \multicolumn{2}{c}{\textbf{Localization}} & \multicolumn{3}{c}{\textbf{Spatial Relation}} & \multicolumn{3}{c}{\textbf{Grounding}} & \\
\textbf{Models} & \textbf{Size} & \textbf{Label} & \textbf{Pos.} & \textbf{Size} & \textbf{Color} & \textbf{Obj.} & \textbf{Text} & \textbf{O-T} & \textbf{O-O} & \textbf{T-T} & \textbf{O-Box} & \textbf{T-PNLS} & \textbf{T-Box} & \multirow{-2}{*}{\textbf{Total}} \\ \hline
\textbf{InternVL2} & 1B & 35 & 29 & 32 & 24 & 28 & 25 & 17 & 27 & 19 & 0 & 1 & 0 & 237 \\
\textbf{Phi-3-Vision} & 4B & 40 & \textbf{34} & {42} & 39 & \textbf{41} & {42}  & 31 & {33} & 42 & 38 & 13 & 2 & 397 \\
\textbf{Monkey-Chat} & 7B & 36 & 22 & 33 & 27 & 26 & 16 & 9 & 18 & 27 & 0 & 0 & 0 & 214 \\
\textbf{Idefics-2} & 8B & 36 & 23 & 36 & 29 & 31 & 27 & 20 & 21 & 33 & 0 & 0 & 0 & 256 \\
\textbf{InternVL2} & 8B & {42} & 30 & \textbf{46} & 44 & 39 & {42} & 27 & {33} & {45} & 15 & 5 & 0 & 368 \\
\textbf{LLaVA 1.5} & 13B & 30 & 10 & 25 & 20 & 32 & 17 & 16 & 24 & 26 & 33  & 0 & 4 & 243 \\
\textbf{LLaVA-NEXT} & 13B  & 36 & 27 & 37 & 33 & 38 & 38 & 23 & 31 & 37 & 39 & 2 & 0 & 335 \\ 
\textbf{LLaVA-NEXT} & 34B & 39 & 27 & 42 & 39 & 39 & 39 & 28 & 31 & \textbf{46} & \textbf{40} & 37 & 5 & 412\\
\textbf{Idefics} & 80B & 0 & 1 & 21 & 20 & 21 & 17 & 20 & 19 & 20 & 0 & 0 & 0 & 139 \\ 
\hline
\hline
\textbf{Qwen-vl-plus}  & - & 38 & 23 & 32 & 35& 26 & 23 & 24 & 23 & 27 & 34 & 22 & 3 & 310 \\
\textbf{Qwen-vl-max} &  - & 39 & 27 & 41 & 36 & 34 & 33 & 26 & 32 & 37 & 24 & 32 & 5 & 366 \\
\textbf{GPT-4V}  & - & 43 & 33 & {43} & 40  & 37 & 38 & 26 & 26 & {45} & 26 & \textbf{48} & 10 & 415 \\
\textbf{GPT-4o}  & - & \textbf{46} & \textbf{34} & {43}  & {41} & {40} & 42 & {34} & \textbf{37} & 40 & 33 & 46 & \textbf{21} & {457} \\ 
\textbf{Claude 3.5 Sonnet} & - & 42 & 31 & 43 & \textbf{46} & 38 & \textbf{45} & \textbf{39} & {36} & \textbf{46} & 39 & 47 & 11 & \textbf{463} \\
\hline
\end{tabular}
}
\vspace{-0.2em}
\caption{ Empirical results of different models on 11 tasks of MMR Benchmark. The blue columns show PNLS scores and the red columns show box matching scores. The upper and lower halves list open-source and proprietary models, respectively. The highest score for each task is highlighted in bold font.}
\label{tab:results}
\vspace{-1em}
\end{table*}

\subsection{Quantitative results}
Table \ref{tab:results} summarizes the performance of eleven models on all tasks, including counting, PNLS, and IoU scores, as introduced in Section \ref{sec:metrics}. The text grounding task output both text and bounding box, thus are evaluated by two metrics. 

We observe that GPT-4o (launched on May 13, 2024) and Claude 3.5 Sonnet (June 20, 2024) demonstrate superior overall performance, as indicated by the total score and the area covered in the radar chart. They generally outperform GPT-4V (March 14, 2023), highlighting the recent progress of proprietary models. However, we find that some open-source models can occasionally outperform GPT-4 models despite their smaller size.

\paragraph{Model Size v.s. Data Quality}
In our experiment, the performance of most models shows a positive correlation with model size. For example, LLaVA-NEXT-34B surpasses LLaVA-NEXT-13B. However, Phi-3-vision demonstrates impressive performance with only 4.2B parameters, surpassing larger models like Qwen-vl-plus and Qwen-vl-max, and rivaling LLaVA-NEXT-34B and GPT-4 models in many tasks. Despite its success, Phi-3-vision has a similar architecture to LLaVA~\citep{liu2024visual}, suggesting that open-source models suffer from data-hungry issues. Thus, high-quality data is more essential than merely scaling up. This finding is further supported by the significant performance gap between Idefics-80B and its smaller successor, Idefics-2-8B.

In contrast to Phi-3-vision's notable performance on multiple-choice and text recognition questions, it performs poorly on text grounding tasks. A possible explanation is that this task demands high instruction-following ability for formatting longer outputs, which might require a larger model size, as we observed only larger models achieve reasonable performance in these tasks.

\begin{figure*}
    \hspace{-0.3em}
    \vspace{-0.1em}
    \captionsetup{type=figure}
    \begin{minipage}[t]{0.5\textwidth}
        \centering
        \includegraphics[width=0.9\linewidth]{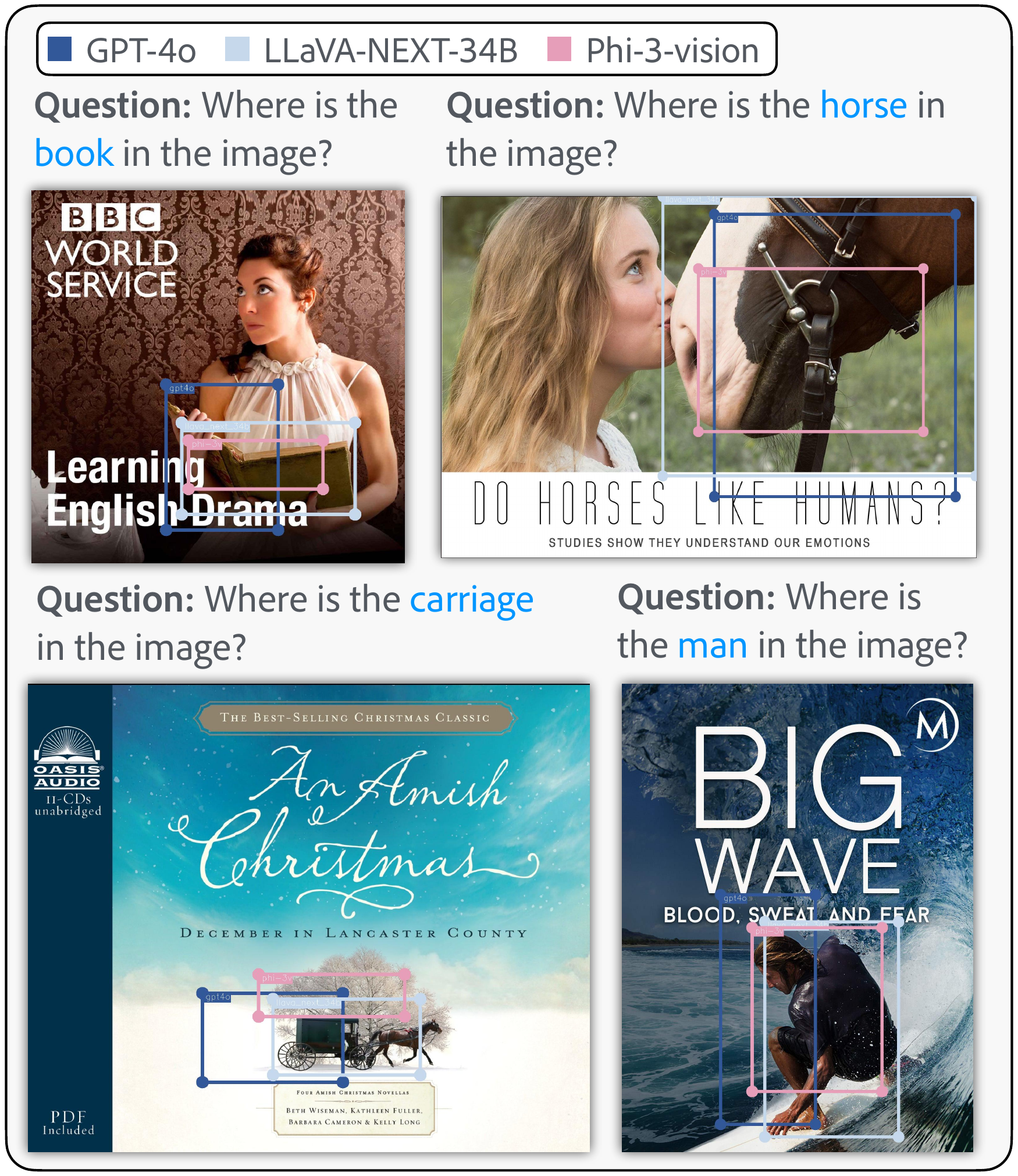}
        \vspace{-0.85em}
        \caption*{(a)}
    \end{minipage}
    \hspace{1em}
    \begin{minipage}[t]{0.5\textwidth}
        \centering
        \includegraphics[width=0.97\linewidth]{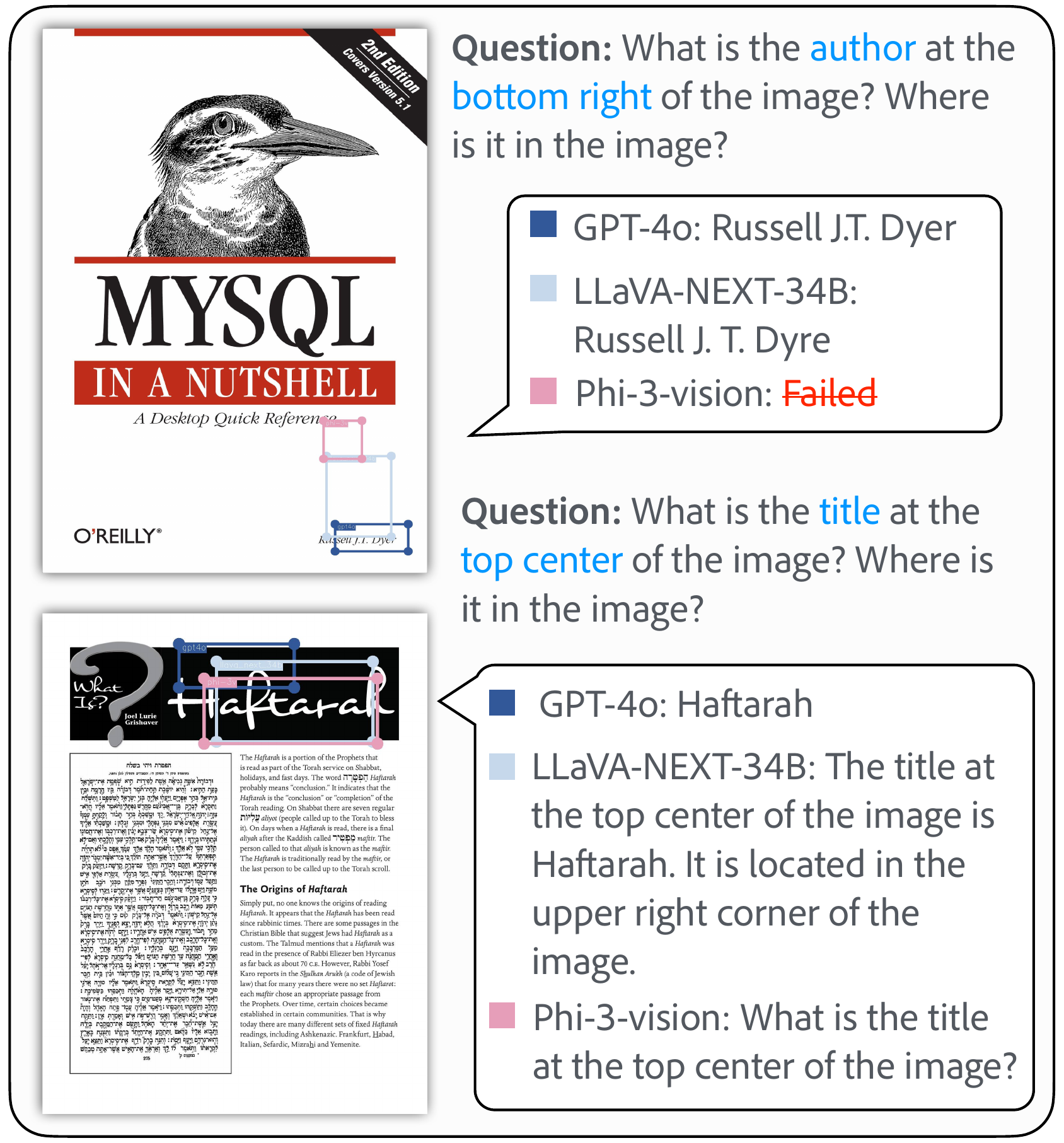}
        \vspace{-0.9em}
        \caption*{(b)}
    \end{minipage}
    \hspace{1em}
    \vspace{-1.0em}
    \caption{Examples generated by different models on the Text (a) and Object (b) Grounding tasks.}
    \vspace{-0.9em}
     \label{fig:examples}
\end{figure*}

\paragraph{Reading ability}
We use PNLS to evaluate the text reading ability of the model. Most models perform well in font recognition and text recognition by label but still struggle to match human performance. Additionally, when text is specified by its rough location, PNLS scores decrease, as these questions are more complex and require spatial understanding before recognition. The task becomes even more challenging with text grounding, where models must output both text and bounding boxes simultaneously. In these cases, smaller models like Idefics-2, LLaVA 1.5, and Monkey-Chat fail to provide valid results. Figure \ref{fig:examples} shows examples of text bounding boxes detected by three models. We can see that LLaVA-NEXT and Phi-3-vision struggle to generate outputs in the required format, and all models, including GPT-4o, are unable to generate accurate bounding boxes. This indicates the need for improved visual text understanding in vision-language models.

\vspace{-0.3em}
\paragraph{Spatial understanding}
We also evaluate the spatial understanding ability of different models in localization, pairwise position understanding, and grounding tasks. Similar to the text grounding results discussed above, some smaller open-source models lack grounding ability and cannot provide valid responses to the questions. However, we find that LLaVA models and Phi-3-vision outperform GPT-4o and significantly outperform GPT-4V in the object grounding task, as measured by bounding box scores and illustrated in Figure \ref{fig:examples}. The excellent performance of LLaVA-NEXT models in these tasks could be attributed to their patch-wise encoding strategy. However, they are trained mainly on natural images with minimal experience in text-rich images~\citep{zhang2023llavar}, resulting in poor performance in text grounding. This highlights the need for annotated text-rich images datasets.

\section{Conclusion}
In this paper, we introduce the Multi-Modal Reading (MMR) benchmark, which evaluates the reasoning and spatial understanding capabilities of LMMs in text-rich image understanding. The benchmark consists of eleven diverse tasks with carefully designed evaluation metrics. The experimental results showcase the performance of different models, giving suggestions on which model to choose in real-world applications. It also underscores the need for further research and development to bridge the gap between LMMs and human-level performance in text-rich image understanding.

\section{Limitations}
We only evaluated recently released models, and more models should be evaluated, which we hope can be handled by the community after the MMR benchmark is released. The evaluation metrics used in MMR still have limitations in accurate evaluation, and we have reformulated the VQA as multiple choices and provided output template for LMMs to alleviate this issue. The questions are all proposed by the GPT-4V which may induce some model bias, while it is still difficult for human annotators to propose suitable questions with complex reasoning as they tend to ask extractive questions. 

\section{Acknowledge}
This work is partially supported by NSF AI Institute-2229873, NSF RI-2223292, an Amazon research award, and an Adobe gift fund. Any opinions, findings and conclusions or recommendations expressed in this material are those of the author(s) and do not necessarily reflect the views of the National Science Foundation, the Institute of Education Sciences, or the U.S. Department of Education.

\section{Ethics Statement}
Multi-Modal Reading (MMR) benchmark adheres to a set of ethical principles and guidelines to ensure responsible and ethical conduct of the study. Informed consent was obtained from all participants, who were fully informed about the purpose and nature of the research. Efforts were made to include participants from diverse backgrounds, promoting inclusivity and representation. The study was conducted with integrity of the research, adhering to scientific rigor and ethical standards. Compliance with relevant laws, regulations, and ethical guidelines was ensured throughout the research process. The research findings aim to contribute to the advancement of AI technology ethically, with a commitment to using the results for the betterment of society.

\appendix
\clearpage
\counterwithin{figure}{section}
\counterwithin{equation}{section}
\counterwithin{table}{section}

\section{Prompt}
\label{appendix:prompt}
This section include all prompt we use for all tasks and models.

\begin{tcolorbox}[colback=LLightGray, colframe=LightGray, title=Object Grounding prompt, boxrule=0pt]
\begin{lstlisting}
where is the {object} in the image?
Please write the position as a
bounding box, and output the 
[x_min, y_min, x_max, y_max] 
coordinates in float numbers in 
python list. Output the text only.
\end{lstlisting}
\end{tcolorbox}

\begin{tcolorbox}[colback=LLightGray, colframe=LightGray, title=Text Grounding prompt, boxrule=0pt]
\begin{lstlisting}
What is the {text label} at the 
{area} of the image? Where is 
it in the image? Please write 
the position as a bounding box, 
and output the [x_min, y_min, 
x_max, y_max] coordinates in float 
numbers in a python list. Output 
the text and bounding box only.
For example: "Hello world" [x_min,
y_min, x_max, y_max]
\end{lstlisting}
\end{tcolorbox}

\begin{tcolorbox}[colback=LLightGray, colframe=LightGray, title=Single Choice prompt, boxrule=0pt]
\begin{lstlisting}
{question} Only print the index of 
the  correct choice as answer, 
such as 1, 2, 3, or 4.
\end{lstlisting}
\end{tcolorbox}

\begin{tcolorbox}[colback=LLightGray, colframe=LightGray, title=Text Recognition prompt, boxrule=0pt]
\begin{lstlisting}
{question} Only print the text; do 
not include any other descriptions.
\end{lstlisting}
\end{tcolorbox}

The prompt is inserted in the required format for each model. For example, LLaVA 1.5 requires the following format:
\begin{tcolorbox}[colback=LLightGray, colframe=LightGray, title=LLaVA 1.5 template, boxrule=0pt]
\begin{lstlisting}
USER: <image>\n<prompt> ASSISTANT:
\end{lstlisting}
\end{tcolorbox}
For the required input format of other models, please refer to the respective source code.

\vfill\eject
\section{PNLS demo}
\label{appendix:pnls}
\begin{figure}[!h]
    \centering
    \includegraphics[width=0.35\textwidth]{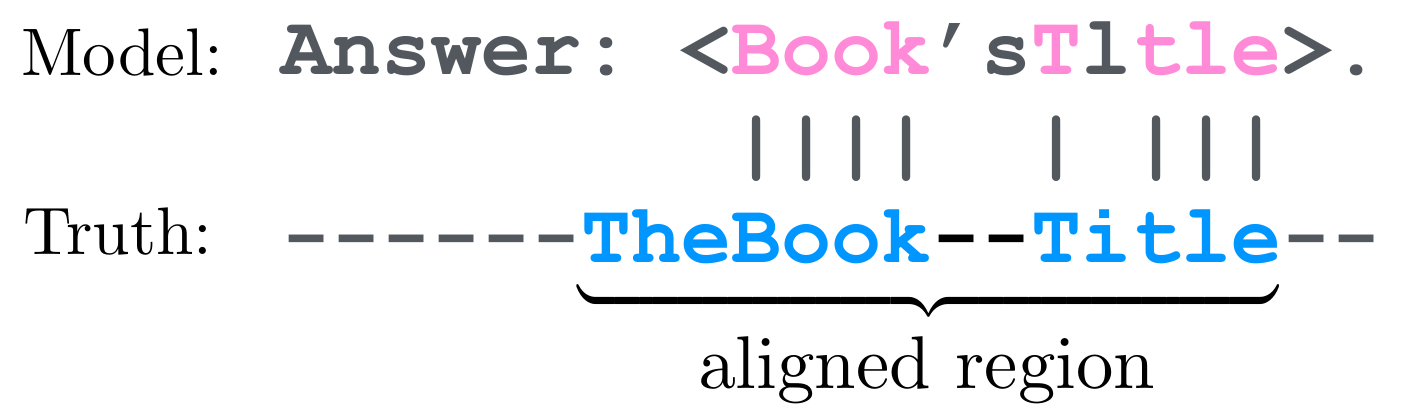}
    \caption{Example of text similarity score. Only the 8 pink characters in the model's output "Answer: <Book'sTltle>." match the true string ("TheBookTitle" in blue). The aligned region has length 14. In this case the similarity is 8/14. }
    \label{fig:text_sim}
\end{figure}

Figure \ref{fig:text_sim} provides an example.

\end{document}